\pdfoutput=1
\documentclass{article} %

\usepackage{iclr2026_conference,times}

\usepackage{latexsym}
\usepackage{color}
\usepackage{palattes}

\usepackage[T1]{fontenc}
\usepackage[utf8]{inputenc}

\usepackage{microtype}
\usepackage{multirow}

\usepackage{inconsolata}
\usepackage{hyperref}
\usepackage{url}
\usepackage{booktabs}
\usepackage{graphicx}
\graphicspath{{figures/}}
\usepackage{longtable}
\usepackage{booktabs}
\usepackage{xspace}
\usepackage{xcolor}
\usepackage{subcaption}
\usepackage{hyperref}
\usepackage{enumitem}
\usepackage{algorithm}
\usepackage{algpseudocode}
\usepackage{amsmath}
\usepackage{float}
\usepackage{array}
\usepackage{caption}
\usepackage{listings}
\usepackage{tcolorbox}
\usepackage{standalone}
\usepackage{tikz}
\usepackage{makecell}
\usepackage{titletoc}

\usepackage{wrapfig}
\usepackage{lineno}

\definecolor{darkblue}{rgb}{0, 0, 0.5}
\hypersetup{colorlinks=true, citecolor=darkblue, linkcolor=darkblue, urlcolor=darkblue}

\lstdefinestyle{json}{
    basicstyle=\ttfamily,
    keywordstyle=\color{blue},
    stringstyle=\color{black},
    backgroundcolor=\color{gray!10},
    frame=single,
    frameround=tttt,
    numbers=none,
    breaklines=true,
    columns=flexible,
    showstringspaces=false
}

\title{Algorithmic Primitives of \\Theory of Mind Benchmarks}

\author{%
  Qiutong Tony Yi$^{*\,}$ \quad
  Wang Bill Zhu$^{*\,}$ \quad
  Robin Jia \quad
  Jesse Thomason \\[0.25em]
  University of Southern California \\[0.25em]
  Corresponding authors: \texttt{qiutongy@usc.edu}, \texttt{wangzhu@berkeley.edu}
}

\usepackage{pifont}
\usepackage{xspace}

\newcommand{\ourmethod}{\textsc{PDDL-Mind}\xspace}
\newcommand{\mmtom}{MMToM-QA\xspace}
\newcommand{\muma}{MuMa-ToM\xspace}
\newcommand{\fantom}{FanToM\xspace}

\iclrfinalcopy %
\begin{document}
\maketitle

\begin{abstract}

Do Theory of Mind (ToM) benchmarks really test theory of mind, or are they actually testing parsing and algorithmic state tracking? While benchmarks vary modality and dialogue length, this paper finds that they re-encode the same few state transition algorithms. We create \ourmethod, a neuro-symbolic approach that recovers such algorithms and upstream inductive biases of ToM benchmarks explicitly by parsing narratives into the Planning Domain Definition Language (PDDL) and executes candidate actions against a predefined set of update algorithms. Doing so provides program-verified state traces to the model. We use our method's performance gain to audit how much of a benchmark's difficulty our inductive bias absorbs. Across \mmtom, \muma, and \fantom, outsourcing parsing to programs alone brings performance close to strong agentic pipelines, while outsourcing state tracking to PDDL moves GPT-4o to close-to-human performance. We demonstrate that complex ToM benchmarks primarily punishes Large Language Models' unreliable implicit implementation of state tracking and parsing algorithms rather than limitations in belief and counterfactual reasoning.
\end{abstract}

\begin{wrapfigure}[20]{r}{0.48\textwidth}
\vspace{-25pt}
\centering
\includegraphics[width=\linewidth]{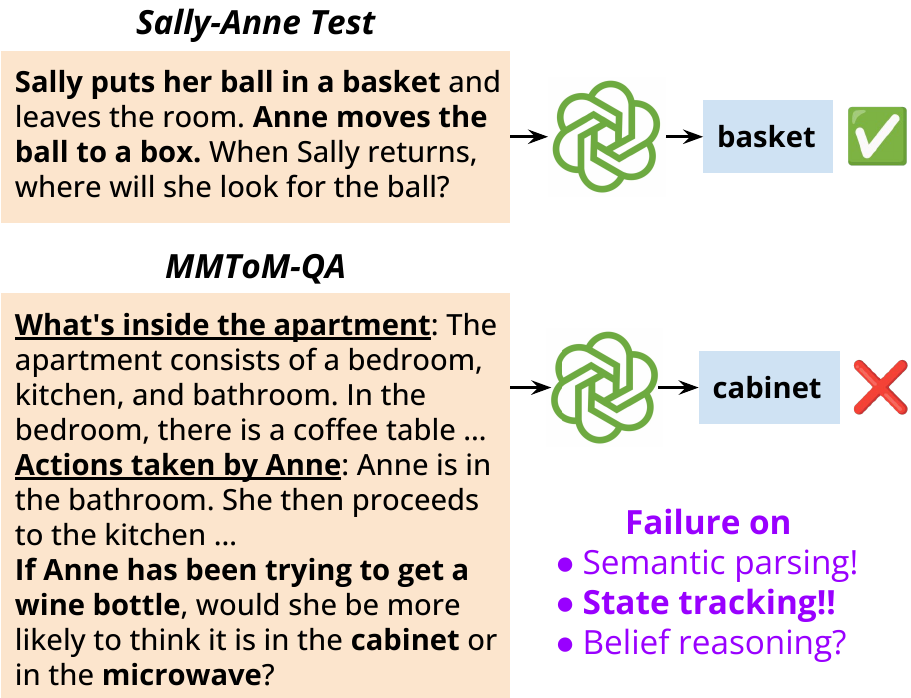}
\caption{Frontier large language models now often pass compact Sally--Anne-style false-belief tests, but they remain below human performance on harder ToM benchmarks. Error attribution becomes opaque when narratives become more complex}
\label{fig:teaser}
\vspace{-5pt}
\end{wrapfigure}

\section{Introduction}
\label{sec:intro}

Theory of Mind (ToM) benchmarks present narratives in which agents act in the world and observers must infer what those agents believe, know, or intend.
These narratives typically take one of two forms: descriptions of \textit{physical actions} in an environment, or \textit{dialogue transcripts} between agents.
These variations share a common underlying structure: each narrative is fundamentally a sequence of events that induces a sequence of world states~\citep{huang-etal-2024-notion, xu-etal-2025-enigmatom}.

The Sally--Anne and related false-belief tasks, illustrated in Figure~\ref{fig:teaser},  have long served as compact tests of whether a reasoner distinguishes the true world state from an agent's belief state~\citep{wimmer1983beliefs, baroncohen1985does}.
Since 2023, researchers have adapted these false-belief tests to large language models (LLMs).
Many LLMs could solve many classic false-belief tasks~\citep{kosinski2024evaluating}, but follow-up studies found that performance could be brittle under small task variations and prompt changes~\citep{ullman2023large, ma2023tomchallenges}.
To test frontier models' belief and counterfactual reasoning ability, recent benchmarks add longer event chains, richer environments, more agents, dialogue, multimodal inputs, and more varied questions.
On these newer tasks, even frontier LLMs remain substantially below human level~\citep{jin2024mmtom, kim2025tracing}.
The prevalent  response has been to scale test-time reasoning: longer chain-of-thought traces, or agentic pipelines that repeatedly refine mental-state hypotheses~\citep{kim2025tracing, zhang2025autotom}.
This buys accuracy with a growing budget of LLM calls, but blurs which capability the extra compute is compensating for.

Complex benchmarks also make failure diagnostics harder.
One hypothesis \cite{huang-etal-2024-notion} attributes errors to \textit{state-history parsing}: the model may fail before reasoning begins because it misreads who is present, which object moved, or which utterance was heard.
A second hypothesis attributes errors to \textit{state tracking}( \cite{sclar-etal-2023-minding-symbolictom, xu-etal-2025-enigmatom}): the model may parse events correctly but fail to maintain a consistent world state as actions accumulate.
A third hypothesis attributes errors to \textit{belief reasoning}: the model may have the right state trace but still fail to infer what a partially informed agent knows, falsely believes, or intends.

These hypotheses are difficult to separate within end-to-end prompting pipelines. We thus introduce \ourmethod as a diagnostic tool. \ourmethod makes the aforementioned event chain assumption explicit with a predefined Planning Domain Definition Language (PDDL; \citealp{aeronautiques1998pddl}) domain. In the domain file, we define modularized functions corresponding to the above 3 hypotheses with strict schema. The domain specifies the benchmark's relevant action vocabulary, preconditions, and postconditions. Given this fixed domain, an LLM parses the initial environment into a PDDL problem file and parses the narrative into a candidate action sequence. PDDL execution then verifies which actions are valid and produces an explicit state trace. Finally, the LLM answers the ToM question using the verified trace.

Under our method, we can then easily infer where the benchmark is putting pressure on the model: If parsing is the main bottleneck, then a model given structured actions but no verified state trace should already reach high performance. If state tracking is the main bottleneck, then PDDL verification should create a large gain. If belief reasoning is the main bottleneck, then large errors should remain even after state tracking is verified. The predefined domain is crucial for this disentanglement for providing isolated capability definitions as helper functions. 

We are able to improve benchmark scores substantially by using \ourmethod to outsource parsing and state-tracking. On \mmtom, \muma, and \fantom, valid parsing alone brings performance within a 10\% difference to several agentic ToM pipelines. Verified state tracking yields another large improvement, moving GPT-4o to near human performance on the first two benchmarks. These gains come with minimal costs. \ourmethod issues exactly \textbf{three LLM calls} per example: one to parse the initial state, one to parse the action sequence, and one to answer the question, while PDDL verification is deterministic symbolic execution with minimal inference cost. 

Our goal is not only to climb scores on benchmarks. We argue that \ourmethod's success should be treated by future benchmark creators as a valid self-diagnostic protocol.
If PDDL verification solves a ToM benchmark, the benchmark may primarily test parsing and state tracking rather instead of belief reasoning.
If errors remain under verified states, those cases are stronger evidence of belief or counterfactual reasoning failures.
This distinction can help benchmark creators design tasks that target the capability they intend to measure.

\section{Related Work}
\label{sec:related}

\paragraph{Classic and LLM false-belief tests.}
The false-belief paradigm asks whether a reasoner can represent that another agent's belief may differ from reality~\citep{wimmer1983beliefs}.
The Sally--Anne task is the canonical compact form of this test~\citep{baroncohen1985does}.
Recent LLM work revived this paradigm: \citet{kosinski2024evaluating} reported that newer LLMs solved many false-belief tasks, while \citet{ullman2023large} and \citet{ma2023tomchallenges} showed that performance remains fragile under small variations and prompt changes.
Our work takes this debate as motivation but focuses on harder modern benchmarks where simple false-belief success is no longer sufficient.

\paragraph{ToM benchmarks.}
Textual and multimodal ToM benchmarks now test longer and more diverse narratives, including ToMi~\citep{le-etal-2019-revisiting-tomi}, BigToM~\citep{gandhi2023understanding}, Hi-ToM~\citep{wu-etal-2023-hitom}, FanToM~\citep{kim2023fantom}, OpenToM~\citep{xu-etal-2024-opentom}, NegotiationToM~\citep{chan-etal-2024-negotiationtom}, \mmtom~\citep{jin2024mmtom}, and \muma~\citep{shi2024mumatommultimodalmultiagenttheory}.
These benchmarks differ in modality, number of agents, and question types, but they commonly present an ordered sequence of events that changes what agents can observe or remember.
This shared structure makes them a natural target for diagnostic state-tracking interventions.
At the same time, benchmark saturation matters for diagnosis.
Prior work reports near-ceiling performance on several older or more templated ToM datasets with chain-of-thought prompting alone~\citep{xu-etal-2025-enigmatom, kim2025tracing}, leaving little headroom for identifying which component fails.
We therefore evaluate on three benchmarks that still exhibit substantial gaps while spanning the principal task categories: \mmtom for single-agent action sequences, \muma for multi-agent action and conversation, and \fantom for multi-agent conversation.
Our goal is to test a failure-mode hypothesis across representative settings, not to exhaustively cover every ToM dataset.

\paragraph{Prompting and agentic ToM methods.}
Prompting-based methods attempt to elicit mental-state reasoning without changing model weights.
SymbolicToM constructs explicit belief graphs for each character~\citep{sclar-etal-2023-minding-symbolictom}; SimToM filters narratives to the events observable to a target agent~\citep{wilf-etal-2024-think-simtom}; TimeToM emphasizes temporal structure~\citep{hou-etal-2024-timetom}.
Agentic and probabilistic methods go further.
BIP-ALM~\citep{jin2024mmtom}, LIMP~\citep{shi2024mumatommultimodalmultiagenttheory}, ThoughtTracing~\citep{kim2025tracing}, and AutoToM~\citep{zhang2025autotom} maintain or search over hypotheses about beliefs, goals, and actions.
These methods often improve final accuracy, but they do so by multiplying the number of LLM calls per example, and the final answer still makes it hard to tell whether success came from better parsing, better state tracking, or better belief reasoning.

\paragraph{Discrete world models and symbolic verification.}
DWM~\citep{huang-etal-2024-notion}, EnigmaToM~\citep{xu-etal-2025-enigmatom}, and Social World Models~\citep{zhou2026socialworldmodels} explicitly model ToM narratives as structured state transitions.
We share their view that ToM benchmarks have event-driven structure.
Social World Models introduce S$^3$AP, a structured social-world representation that makes agents, actions, observations, and mental states explicit.
The difference is that \ourmethod fixes the transition semantics in a predefined PDDL domain instead of asking the LLM to infer them or represent them in free-form structure.
This connects our method to work showing that language models can be useful interfaces to symbolic planning languages~\citep{llmp, zhu-etal-2025-language-models}, while also reflecting a competence-aligned scaffolding principle: use the LLM where it is reliable, and externalize the part that needs exact verification~\citep{agrawal2025gepa}.

\section{\ourmethod}
\label{sec:method}

\begin{figure*}[t]
\centering
\includegraphics[width=\textwidth]{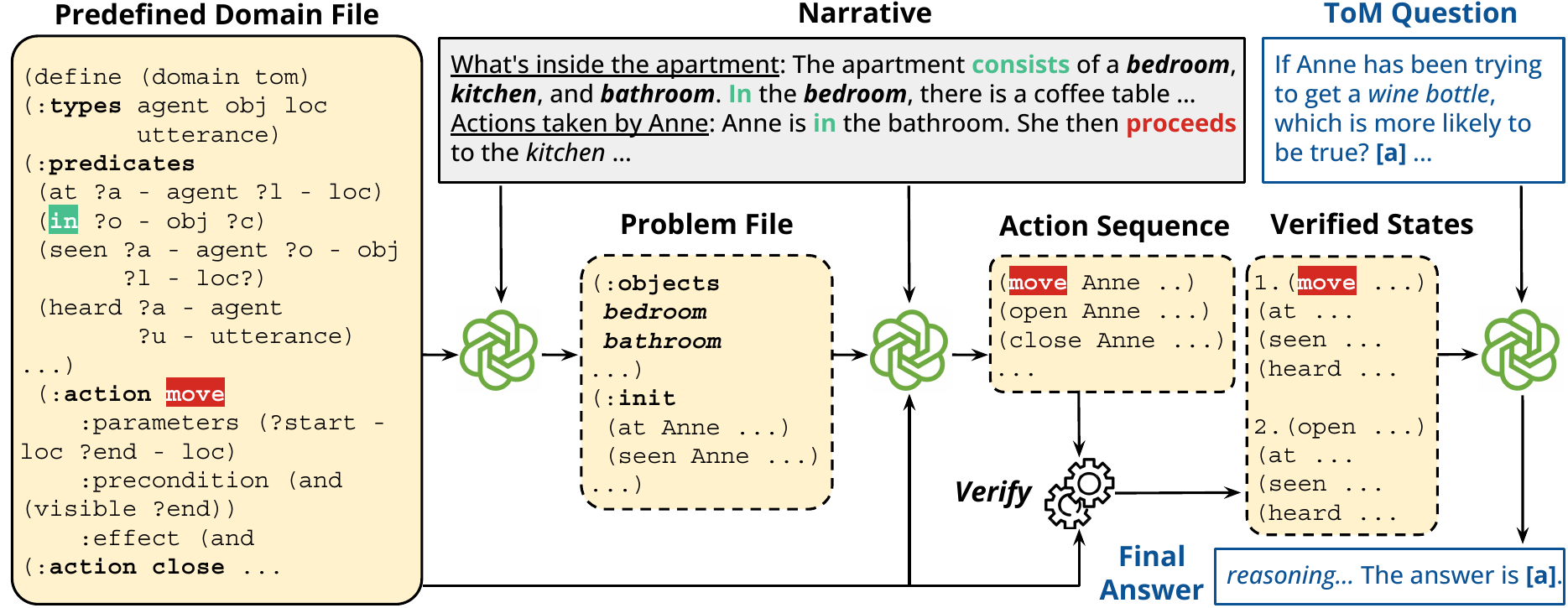}
\caption{Starting from a predefined PDDL domain file, \ourmethod parses the narrative into an initial world state and a sequence of candidate actions. PDDL verification executes only actions whose preconditions hold, producing a reliable state trace for final ToM question answering.}
\label{fig:method}
\end{figure*}

We introduce \ourmethod, a neuro-symbolic framework that decouples environment state evolution from belief inference (Figure~\ref{fig:method}).
The framework is designed as a diagnostic instrument: each stage corresponds to one hypothesized failure mode in ToM evaluation.

\paragraph{Predefined domain file.}
We begin with a predefined PDDL domain file.
The domain is intentionally small because it is meant to capture the transition types that recur across current ToM benchmarks, not to model every detail of a possible world.
Physical-action benchmarks require agents, locations, objects, containers, visibility, and object manipulation.
Dialogue benchmarks require agents, utterances, conversational presence, and memory of heard utterances.

We therefore define four object types: \texttt{agent}, representing people mentioned in the narrative; \texttt{loc}, representing locations; \texttt{obj}, representing physical items; and \texttt{utterance}, representing spoken messages.
We define predicates such as \texttt{at}, \texttt{in}, and \texttt{holding} for physical state.
We also define \texttt{seen}, which records whether an agent has observed an object, and \texttt{heard}, which records whether an agent has heard an utterance.
Finally, we define seven actions: \texttt{move}, \texttt{open}, \texttt{close}, \texttt{grab}, \texttt{drop}, \texttt{ask}, and \texttt{tell}.
Each action has explicit parameters, preconditions, and postconditions.
The complete domain file is provided in Appendix~\ref{appsec:prompt}.

We choose a predefined domain for two reasons.
First, it makes the benchmark's world model inspectable: a reader can see exactly which transition rules are being assumed.
Second, it prevents the diagnostic from mixing two questions that should be separated: whether the LLM can parse a narrative into a known formal schema, and whether it can invent a correct formal schema in the first place.
The latter is an important planning problem, but it is not the same as ToM belief reasoning.

\paragraph{State and action extraction.}
Given the domain file, we prompt an LLM to parse the narrative into a PDDL problem file.
The problem file specifies the objects present in the scenario and the initial state $s_0$.
We define a state $s$ as a set of grounded predicates, such as \texttt{(at Mary bathroom)}.
This stage tests environment parsing: the model must identify the entities and initial relations that the story assumes.

The LLM then parses the narrative into a sequence of candidate actions $(a_1, \dots, a_t)$.
This stage tests event parsing: the model must map language such as ``Mary walks to the kitchen'' or ``John says that the key is in the drawer'' into grounded PDDL actions.
Because narratives may include events outside the domain, we allow the candidate sequence to contain invalid or irrelevant actions.

\paragraph{Verified state tracking.}
For each candidate action $a_i$, we check whether the current state $s_{i-1}$ satisfies the action's preconditions.
If the preconditions hold, we execute the action's postconditions and produce $s_i$.
If the preconditions fail, the action is ignored and the state remains unchanged.
This gives us a verified sequence of actions and world states.
Verification is deterministic symbolic execution: it involves no LLM inference and its compute cost is negligible next to a single LLM call.
The verification stage tests the state-tracking hypothesis: if state tracking is the bottleneck, providing the verified state trace should substantially improve ToM question answering.

\paragraph{ToM question answering.}
Finally, the LLM receives the verified state sequence and a ToM question, and answers with chain-of-thought reasoning.
At this point, the model no longer needs to reconstruct the hidden world state from the original prose.
Residual errors therefore provide a cleaner signal about belief reasoning, counterfactual reasoning, or answer selection.
In total, \ourmethod issues three LLM calls per example: problem-file parsing, action parsing, and question answering, with all state tracking delegated to the symbolic executor.

\paragraph{Diagnostic variants.}
We evaluate two ablations to isolate the role of each component.
\ourmethod without verification gives the model the parsed symbolic events but does not provide the verified intermediate states; this measures how far valid parsing goes without reliable state execution.
\ourmethod without a predefined domain asks the model to construct its own domain before parsing; this measures whether failures come from inducing the world model rather than using it.
Together with the full method, these variants form the diagnostic ladder used in Section~\ref{sec:experiments}.

\section{Experiments and Results}
\label{sec:experiments}

\begin{table*}[t]
\centering
\small
\tabcolsep 12pt
\begin{tabular}{lrrr}
\toprule
\bf Method & \bf \mmtom & \bf \muma & \bf \fantom \\
\midrule
BIP-ALM$^{\star}$ & 56.2 & 33.9 & - \\
LIMP$^{\star}$ & 55.3 & 76.6 & - \\
TT$^{\star}$ & 69.0 & - & 54.7 \\
AutoToM$^{\star}$ & 83.0 & 81.4 & - \\
EnigmaToM$^{\star}$ & - & - & 55.3 \\
S$^3$AP$^{\star}$ & 69.2 & - & 49.1 \\
\midrule
Human$^{\star}$ & 82.5 & 93.5 & 87.5 \\
\midrule
\ourmethod w/o predefined domain & 74.9 & 78.9 & 69.7 \\
\ourmethod w/o verification & 77.4 & 79.6 & 74.7 \\
\ourmethod & \bf 88.3 & \bf 88.8 & \bf 80.0 \\
\bottomrule
\end{tabular}
\caption{Main diagnostic results. The no-domain ablation tests whether LLMs can induce the transition rules; the no-verification ablation tests structured parsing without verified state execution. All \ourmethod variants use three LLM calls per example. $^{\star}$ denotes results from previous papers. The S$^3$AP row reports GPT-4o results with the primary o3 parser from Social World Models.}
\label{tab:main_diagnostic}
\end{table*}

We evaluate \ourmethod on three representative ToM benchmarks.
\mmtom~\citep{jin2024mmtom} contains 600 questions about belief and goal inference from long household action sequences.
\muma~\citep{shi2024mumatommultimodalmultiagenttheory} contains 900 multi-agent action and conversation examples.
\fantom~\citep{kim2023fantom} contains information-asymmetric conversations with questions about answerability and accessibility.
All tasks are evaluated as multiple-choice question answering by accuracy.
We choose these three benchmarks because they cover the main narrative forms we want to diagnose: single-agent action sequences, mixed multi-agent action and conversation, and multi-agent conversation.
We do not treat coverage as exhaustive.
Datasets such as ToMi and BigToM remain important, but prior work reports that several older or more templated ToM settings are close to saturated under chain-of-thought prompting~\citep{xu-etal-2025-enigmatom, kim2025tracing}.
Such saturation leaves little headroom for diagnosing via ablations. We therefore focus on benchmarks that still show substantial performance gaps while spanning action-only, mixed, and conversational ToM settings.

We compare against BIP-ALM~\citep{jin2024mmtom} and LIMP~\citep{shi2024mumatommultimodalmultiagenttheory}, which combine Bayesian inverse planning with language models; ThoughtTracing (TT; \citealp{kim2025tracing}), which generates and weights hypotheses about mental states; AutoToM~\citep{zhang2025autotom}, which iteratively constructs agent models for ToM inference; EnigmaToM~\citep{xu-etal-2025-enigmatom}, which tracks events with a neural knowledge base; and S$^3$AP~\citep{zhou2026socialworldmodels}, which parses narratives into structured social-world state representations before answering.
We use GPT-4o~\citep{openai2024gpt4ocard} for the main \ourmethod results. 

\paragraph{Parsing helps, but does not solve ToM.}
Table~\ref{tab:main_diagnostic} merges the main result table with the two diagnostic ablations.
\ourmethod without verification reaches $77.4\%$ on \mmtom, $79.6\%$ on \muma, and $74.7\%$ on \fantom.
This is already close to several agentic pipelines, despite using only structured parsing and no verified state trace.
The result supports the first hypothesis: a large part of the improvement available to ToM systems comes from turning event narratives into a cleaner representation.

\paragraph{Verified state tracking closes most of the gap at no extra inference cost.}
Full \ourmethod reaches $88.3\%$ on \mmtom, $88.8\%$ on \muma, and $80.0\%$ on \fantom.
This moves performance near human accuracy on \mmtom and \muma and exceeds the strongest prior method reported for each benchmark in Table~\ref{tab:main_diagnostic}.
The jump from parsed actions to verified states is especially diagnostic: the model is not merely benefiting from a more structured prompt, but from an executable state trace whose transitions have been checked against preconditions and postconditions.
Notably, this jump is free at inference time: verification is deterministic symbolic execution, so the gain over the no-verification variant costs zero additional LLM calls.

\paragraph{\fantom's All metric reveals a separate consistency capability.}
On \fantom, the $80.0\%$ we report is a per-question average over the Accessibility and Answerability subtypes.
The benchmark also defines an All metric that requires a model to answer \emph{every} question type about the same conversation correctly before that context counts as solved~\citep{kim2023fantom}.
The All metric is stricter than the per-question average because it tests answer consistency across question types in addition to per-question correctness: a model that gets each question right in isolation but flips between accessibility and answerability on the same context fails the All metric while scoring well on the average.
Humans reach $87.5\%$ on the per-question average but score noticeably lower under All, and the residual gap between \ourmethod and human accuracy on \fantom should be read in this light.
Reliable state tracking removes most of the per-question gap, but the All metric measures a different capability: maintaining a coherent answer policy across multiple questions about the same mental-state configuration.
This is closer to belief-reasoning consistency than to state tracking, and is precisely the kind of residual capability that our diagnostic framing aims to expose.

\paragraph{Inducing the world model is a separate bottleneck.}
\ourmethod without a predefined domain performs worse than the no-verification variant on all three benchmarks.\footnotetext[1]{FanToM Avg.\ is computed over BeliefQ, AnwerabilityQ, InfoAccessQ.}
This means that asking the LLM to invent the PDDL transition system is harder than asking it to parse into a fixed transition system.
We interpret this as a world-model induction failure, not as direct evidence of a ToM reasoning failure.
For benchmark analysis, this distinction matters: a task may be hard because the relevant transition rules are underspecified or hard to infer, even if belief reasoning over a correct trace would be easy.

\begin{table*}[t]
\centering
\tabcolsep 3pt
\resizebox{\linewidth}{!}{
\begin{tabular}{@{}llccccccccccc@{}}
\toprule
\bf Model & \bf Method & \bf \# Calls & \multicolumn{3}{c}{\bf \mmtom} & \multicolumn{4}{c}{\bf \muma}& \multicolumn{3}{c}{\bf \fantom}\\
\cmidrule(lr){4-6}\cmidrule(lr){7-10} \cmidrule(lr){11-13}
& & & Avg. & Belief & Goal & Avg. & Belief & Goal(S) & Goal(B) & Avg.\footnotemark[1] & Access & Answer \\
\midrule
\multirow{3}{*}{Kimi-K2-Instruct} & CoT & 1 & 54.2 & 70.6 & 42.0 & 58.9 & 63.3 & 66.7 & 48.0 & 46.4 & 43.8 & 26.1\\
& DWM & 4 & 31.7 & 35.3 & 29.0 & 58.5 & 58.9 & 57.8 & 58.9 & 69.8 & 73.4 & 61.6\\
& \ourmethod & 3 & \bf 74.5 & \bf 80.3 & \bf 68.7 & \bf 78.0 & \bf 67.0 & \bf 85.2 & \bf 81.8 & \bf80.0 & \bf73.4 & \bf77.8  \\
\midrule
\multirow{3}{*}{GPT-4o} & CoT & 1 & 61.4 & 85.3 & 43.5 & 41.8 & 57.8 & 31.1 & 37.0 & 45.3 & 27.1 & 25.9\\
& DWM & 4 & 36.0 & 46.0 & 28.5 & 63.2 & 76.7 & 53.3 & 60.0 & 70.9 & \bf78.5 & 70.6 \\
& \ourmethod & 3 & \bf 88.3 & \bf 91.0 & \bf 85.7 & \bf 88.8 & \bf 88.7 & \bf 92.0 & \bf 83.3 & \bf 78.5 &  72.9 & \bf 76.6 \\
\midrule
\multirow{3}{*}{Claude-Sonnet-4.5} & CoT & 1 & 69.4 & 93.3 & 51.5 & 58.6 & 66.7 & 44.4 & 64.0 & 40.5 & 27.1 & 26.0 \\
& DWM & 4 & 66.6 & 76.6 & 59.0 & 86.1 & \bf 98.9 & 72.2 & 87.0 & 61.9 & 56.7 & 55.1\\
& \ourmethod & 3 & \bf 90.9 & \bf 92.7 & \bf 89.0 & \bf 88.6 & 95.6 & \bf 82.2 & \bf 87.5  & \bf79.3 & \bf73.2 & \bf73.8\\
\bottomrule
\end{tabular}
}
\caption{\ourmethod vs.\ cost-efficient methods. \ourmethod consistently outperforms chain-of-thought and DWM on nearly every sub-category of \mmtom, \muma, and \fantom across models, while using fewer LLM calls per example than DWM.}
\label{tab:breakdown}
\end{table*}

\paragraph{Manual analysis separates parsing/state errors from reasoning errors.}
We manually inspect 20 failure cases on each of \mmtom and \muma for the full GPT-4o pipeline.
Following the diagnostic framing, we merge parsing and state-tracking failures into one category: these are cases where the final answer was wrong because the symbolic trace was wrong or incomplete.
The second category contains belief and counterfactual reasoning failures given a correct trace.
Out of these 40 inspected failures, 13 cases ($33\%$) are parsing/state-tracking errors and 27 cases ($67\%$) are belief or counterfactual reasoning errors.
Because the total error rate of full \ourmethod is about $12\%$, the reasoning-only gap after perfect parsing and tracking would be roughly $8\%$ on these benchmarks.

The two failure modes have distinct signatures.
A representative parsing failure on \mmtom (Listing~\ref{lst:fail_parsing}) involves a narrative in which David moves to the fourth kitchen cabinet, opens it, closes it, and then begins opening the third cabinet.
The parser misclassifies \texttt{cabinet\_kitchen\_4} as a location rather than an object, asserting \texttt{(is-loc cabinet\_kitchen\_4)} in the initial state.
This violates the preconditions of \texttt{open} on a container, so the verifier silently halts the trace after the \texttt{move} step.
The downstream model then answers the goal-inference question from an impoverished state in which no cabinets are recorded as seen.
A representative reasoning failure (Listing~\ref{lst:fail_reasoning}) shows the opposite pattern: the trace is correct --- Karen opens the fridge and microwave, and the verifier correctly registers the chips revealed inside the microwave --- but the model infers from ``saw chips but did not take them'' that Karen \emph{was} searching for chips, when the standard goal-inference reading is that an agent who passes over a visible object is unlikely to want it.
Together, these examples illustrate the two error modes whose ratio our diagnostic aims to isolate.
Further qualitative cases are provided in Appendix~\ref{appsec:error}.

\paragraph{Generalizability across models and categories.}
Table~\ref{tab:breakdown} compares \ourmethod with cost-efficient baselines across model backbones and question categories.
\ourmethod consistently improves over chain-of-thought and DWM across \mmtom, \muma, and \fantom.
The gains hold for Kimi-K2-Instruct, GPT-4o, and Claude-Sonnet-4.5, although weaker models show more parsing errors and stronger models leave less room for improvement in saturated categories.
The main exception is \muma belief questions with Claude-Sonnet-4.5, where both DWM and \ourmethod exceed $95\%$ accuracy and DWM is modestly higher.
At that level, a small number of parsing errors can dominate the residual difference.

\paragraph{\ourmethod is efficient as well as accurate.}
\ourmethod uses three LLM calls per example: one for the initial state, one for the action sequence, and one for final ToM inference.
This is fewer than agentic methods that repeatedly resample, score, or refine hypotheses: AutoToM, the strongest prior method on \mmtom and \muma, reports an average of 17.6 LLM calls per example on \mmtom and 35.1 on \muma, so \ourmethod exceeds its accuracy with roughly $6\times$ and $12\times$ fewer calls.
Among methods with comparably small, fixed budgets (Table~\ref{tab:breakdown}), \ourmethod also dominates: it outperforms DWM on nearly every sub-category while using three calls to DWM's four.
The diagnostic gain therefore does not come from spending more inference-time compute; it comes from moving exact state execution out of the LLM and into a symbolic verifier that costs nothing to run.

\section{Why PDDL Works as a Diagnostic}
\label{sec:why_pddl}

The diagnostic strength of \ourmethod rests on a structural property shared by current ToM benchmarks: their narratives, regardless of surface form, reduce to a sequence of typed events that update a small, closed-world state.
This section makes that property explicit and explains why PDDL provides the right level of abstraction to expose it.

\subsection{A Unified Event--State View of ToM Benchmarks}
\label{sec:structural_abstraction}

\begin{figure}[ht]
    \centering
    \captionsetup[subfigure]{skip=2pt}
    \begin{subfigure}[b]{0.38\linewidth}
        \centering
        \resizebox{\linewidth}{!}{\centering
\begin{tikzpicture}[
    every node/.style={circle, fill=black, inner sep=2pt},
    label/.style={circle, fill=none, inner sep=0pt, font=\small}
]

\def\rowone{0}
\def\rowtwo{-1.5}
\def\rowthree{-3}
\def\rowfour{-4.5}
\def\rowfive{-6}

\def\colone{0}
\def\coltwo{1.5}
\def\colthree{3}
\def\colfour{4.5}

\begin{scope}[
    xshift=-5.5cm,
    yshift=-1.5cm,
    level distance=1.5cm,
    level 1/.style={sibling distance=2.5cm},
    level 2/.style={sibling distance=1.5cm}
]
\node[darkgray] (root) {}
    child {
        node[darkgray] (L) {}
        child { node[darkgray] (LL) {} }
        child { node[darkgray] (LR) {} }
    }
    child {
        node[darkgray] (R) {}
        child { node[darkgray] (RL) {} }
        child { node[darkgray] (RR) {} }
    };
\draw[->, line width=1.75pt, p8e, opacity=1] (root) to[bend right=55] (L);
\draw[->, line width=1.75pt, p3j, opacity=1] (L) to[bend left=60] (LR);
\draw[->, line width=1.75pt, p3g, opacity=1] (R) to[bend left=60] (RL);
\draw[->, line width=1.75pt, p3g, opacity=1] (RR) to[bend left=60] (R);
\draw[->, line width=1.75pt, p8e, opacity=1] (RL) to[bend left=60] (R);
\draw[->, line width=1.75pt, p9g, opacity=1] (R) to[bend left=60] (RR);
\draw[->, line width=1.75pt, p8e, opacity=1] (R) to[bend right=60] (root);
\end{scope}

\begin{scope}[xshift=0.5cm, yshift=0cm]
    \foreach \i in {0,...,4} {
        \node (v\i) at (-1.5, -\i*1.5) {};
    }
    \draw[p9g,line width=1.75pt] (v0) -- (v1);
    \draw[p3g,line width=1.75pt] (v1) -- (v2);
    \draw[p8e,line width=1.75pt] (v2) -- (v3);
    \draw[p3j,line width=1.75pt] (v3) -- (v4);
\end{scope}

\begin{scope}[xshift=1.5cm, yshift=0cm]
    \def\rowone{0}
    \def\rowtwo{-1.5}
    \def\rowthree{-3}
    \def\rowfour{-4.5}
    \def\rowfive{-6}
    \def\colone{0}
    \def\coltwo{1.5}
    \def\colthree{3}
    \def\colfour{4.5}

    \node[p9g] (r1p1) at (\colone, \rowone) {};
    \node[p3g] (r1p2) at (\coltwo, \rowone) {};
    \node[p3j] (r1p3) at (\colthree, \rowone) {};
    
    \node[p9g] (r2p1) at (\colone, \rowtwo) {};
    \node[p3g] (r2p2) at (\coltwo, \rowtwo) {};
    \draw[dashed,p3g,line width=1.75pt](r2p2) circle (0.25cm);
    \node[p3j] (r2p3) at (\colthree, \rowtwo) {};
    
    \node[p9g] (r3p1) at (\colone, \rowthree) {};
    \node[p3g] (r3p2) at (\coltwo, \rowthree) {};
    \node[p3j] (r3p3) at (\colthree, \rowthree) {};
    \node[p8e] (r3p4) at (\colfour, \rowthree) {};
    \draw[dashed,p8e,line width=1.75pt](r3p4) circle (0.25cm);
    
    \node[p9g] (r4p1) at (\colone, \rowfour) {};
    \node[p3j] (r4p3) at (\colthree, \rowfour) {};
    \node[p8e] (r4p4) at (\colfour, \rowfour) {};
    
    \draw[p9g,line width=1.75pt] (r1p1) -- (r2p1);
    \draw[p9g,line width=1.75pt] (r1p1) -- (r2p2);
    \draw[p9g,line width=1.75pt] (r1p1) -- (r2p3);
    
    \draw[p3g,line width=1.75pt] (r2p2) -- (r3p1);
    \draw[p3g,line width=1.75pt] (r2p2) -- (r3p2);
    \draw[p3g,line width=1.75pt] (r2p2) -- (r3p3);
    
    \draw[p8e,line width=1.75pt] (r3p4) -- (r4p1);
    \draw[p8e,line width=1.75pt] (r3p4) -- (r4p3);
    \draw[p8e,line width=1.75pt] (r3p4) -- (r4p4);
    
    \node[p9g] (r5p1) at (\colone, \rowfive) {};
    \node[p3j] (r5p3) at (\colthree, \rowfive) {};
    \node[p8e] (r5p4) at (\colfour, \rowfive) {};
    
    \draw[p3j,line width=1.75pt] (r4p3) -- (r5p1);
    \draw[p3j,line width=1.75pt] (r4p3) -- (r5p3);
    \draw[p3j,line width=1.75pt] (r4p3) -- (r5p4);
\end{scope}

\draw[dashed, gray ,opacity=0.7] (-2.5, 1) -- (-2.5, -7);
\draw[dashed, gray, opacity=0.7] (0.5, 1) -- (0.5, -7);

\end{tikzpicture}%
}
        \caption{}
        \label{fig:structural-views}
    \end{subfigure}
    \hfill
    \begin{subfigure}[b]{0.535\linewidth}
        \centering
        \includegraphics[width=\linewidth, trim=148 214 194 218, clip]{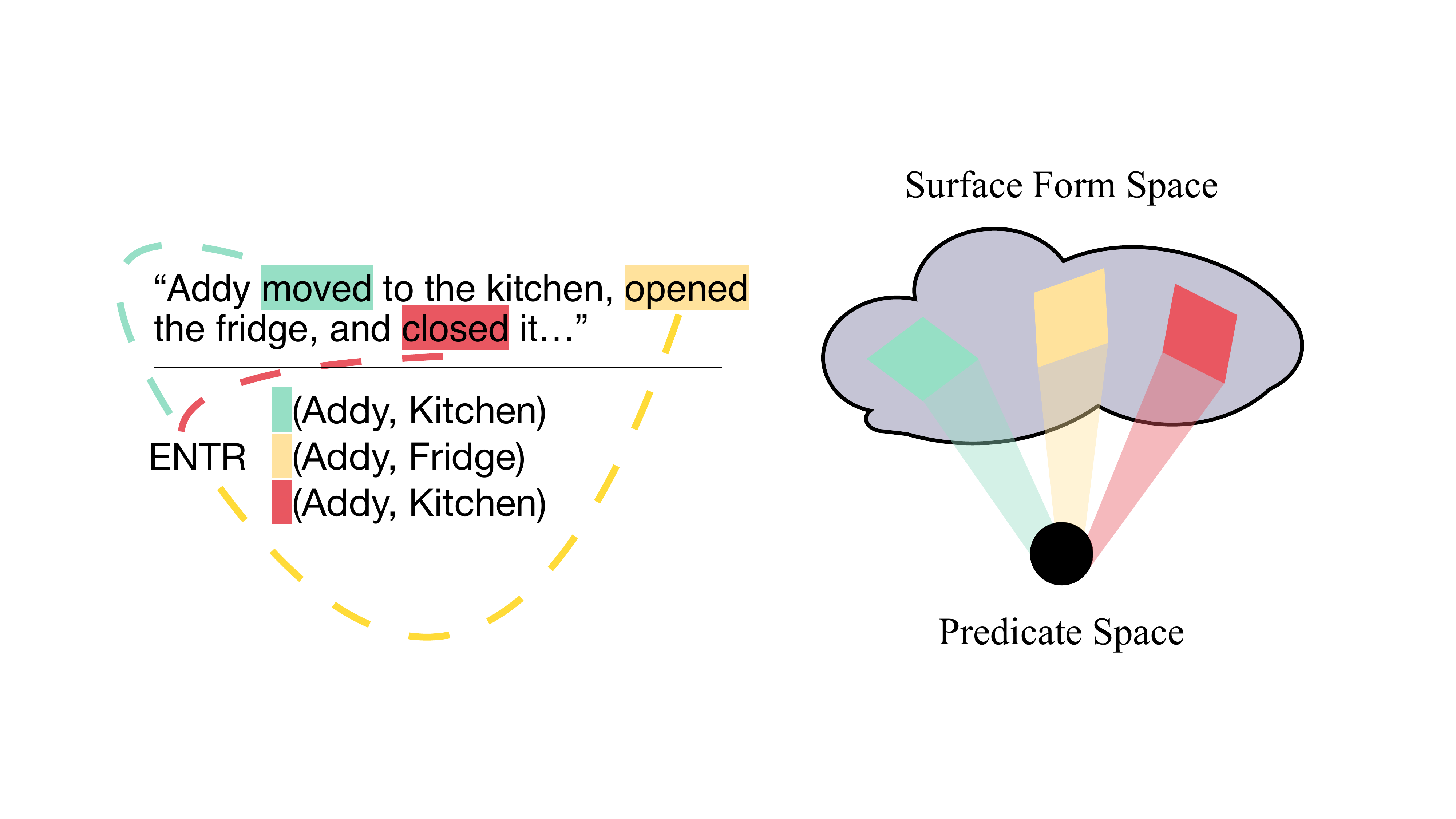}
        \caption{}
        \label{fig:surface-predicate}
    \end{subfigure}
    \caption{(a) Three structural views of ToM benchmark instances, with colors denoting timesteps.
    Left: single-agent action sequences (e.g., \mmtom) induce a hierarchical containment tree; each action updates the agent's position or perceptual access in this tree.
    Middle: the unified event--state abstraction shared across benchmarks --- a chain of states connected by typed events with preconditions and effects.
    Right: multi-agent conversation (e.g., \fantom) induces broadcast edges from a speaker to all present listeners, with join/leave events modifying the recipient set.
    (b) Different surface verbs (moved, opened, closed) map to one helper function (ENTR) in predicate space.}
    \label{fig:world-modeling}
\end{figure}

Figure~\ref{fig:structural-views} summarizes this structure through three complementary views.
The left panel illustrates a Sally--Anne-style single-agent scenario: rooms contain containers, containers contain objects, and each action updates either the agent's location in this containment tree or their perceptual access (e.g., opening a container).
State tracking here reduces to maintaining the agent's position and the visibility predicate at each node, which is computationally a tree traversal, a stricter problem formation than \citep{xu-etal-2025-enigmatom}.
The right panel illustrates multi-agent conversation as broadcast within a shared environment: when an agent speaks, all currently present agents store the utterance, while join and leave events modify the set of recipients.

State tracking here is acyclic information flow over a dynamically changing set of agents, expressible through predicate updates such as \textsc{Speak}, \textsc{Hear}, \textsc{Present}, \textsc{Join}, and \textsc{Leave}.

The middle panel captures what these two regimes share: a chain of states connected by typed events, read from top to bottom.Both physical actions and conversational updates are events with preconditions (who must be present, what must be true) and effects (what predicates change). PDDL is exactly this abstraction. This should be read together with (b): another interpretation of why \ourmethod is successful is because we identified shared algorithmic structure in distinct surface forms in the ToM stories. Which is why LLMs, during parsing, bin different verbs to the same PDDL action. By committing to a predefined PDDL domain, \ourmethod inherits a single representation that covers single-agent action sequences, multi-agent action sequences, and multi-agent conversation.

\subsection{Event Structure Enables Reliable Parsing}
\label{sec:event_structure}

The first reason PDDL helps follows from the structure above but is not unique to PDDL.
Because each instance describes a sequence of actions, utterances, entrances, exits, observations, or object movements, the parsing problem is unusually favorable for LLMs.
The model only needs to identify event arguments and map them into a small number of transition types.

This also explains why other agentic frameworks can perform well.
DWM, EnigmaToM, ThoughtTracing, and AutoToM all exploit the same event-chain structure even when they represent the intermediate state differently.
At the parsing level, \ourmethod without verification is therefore close to several agentic pipelines.
The shared lesson is that ToM benchmark narratives are not unstructured prose: they already contain a latent algorithmic trace.

However, parsing is only the first half of the problem.
A parsed event sequence can still be inconsistent, incomplete, or subtly wrong.
An action may refer to an object that is not visible, an agent may hear an utterance only if present, and opening a container may reveal new objects only under specific spatial conditions.
These constraints are easy to violate when state tracking remains implicit in the language model's context.

\subsection{Preconditions and Effects Verify State Tracking}
\label{sec:verification_benefit}

The second reason PDDL helps is specific to symbolic execution.
PDDL actions are transition functions with explicit preconditions and effects: preconditions say when an event can legally apply, and effects say exactly how the world state changes afterward.
This lets \ourmethod check each candidate event before committing it to the state trace.

This verification step is the crucial diagnostic advantage over purely neural state trackers \citep{xu-etal-2025-enigmatom}.
If a parser proposes an impossible action, PDDL execution rejects it rather than letting the error silently contaminate later reasoning.
If an action is valid, its effects are applied deterministically, so the resulting state trace is internally consistent.
The model answering the final ToM question is therefore conditioned on a trace whose state transitions have been verified rather than merely narrated.

The predefined domain is what makes this possible.
Without it, the model must infer not only which event occurred, but also what that event means as a transition rule.
Our no-domain ablation shows that this world-model induction step is unreliable.
That failure is scientifically interesting, but it is a different target from belief reasoning.
PDDL gives benchmark designers a way to hold the transition rules fixed and ask: once the world state is reliable, does the model still fail at ToM?

\section{Discussion and Conclusion}
\label{sec:discussion}

Our results suggest a reinterpretation of what current ToM benchmarks actually measure.
The headline is not that \ourmethod is another SOTA method. We also want to emphasize that the gap between end-to-end LLM accuracy and human accuracy decomposes cleanly into three layers: 
Parsing the narrative into a structured event sequence accounts for a substantial share of the gain.
Verified state tracking accounts for most of what remains.
The residual belief-reasoning gap, isolated by holding the world state fixed and reliable, is much smaller than the previous end-to-end gap had suggested.
The decomposition also carries an efficiency lesson: by precisely isolating the real missing capability---bookkeeping rather than inference---the fix is nearly free. Three LLM calls plus a deterministic symbolic executor outperform agentic pipelines that spend an order of magnitude more test-time compute.

We wish \ourmethod can serve as a measuring instrument: first, write or adapt a PDDL domain that captures the intended event semantics.
Second, run the benchmark under the three diagnostic settings: no predefined domain, no verification, and full verification.
Third, inspect the failures that remain under full verification.
If the full method nearly solves the benchmark, the benchmark is mostly testing parsing and state tracking.
If many failures remain, those examples are stronger candidates for testing belief, counterfactual, or intention reasoning.

Benchmarks aimed at belief reasoning should should further challenge LLMs, creating tasks where models' performance remain low \emph{after} the state trace is made available. Benchmarks aimed at world-model induction should make that target explicit and evaluate whether models can infer the transition rules themselves, rather than bundling rule induction together with belief inference.

By making the hidden algorithmic assumptions of ToM benchmarks visible, executable, and falsifiable, \ourmethod helps separate the capabilities a benchmark currently rewards from the capabilities it was designed to probe.

Disentangling state tracking from belief inference admits further philosophical discussions illuminating for future benchmark design. Complexity frameworks such as \cite{huang-etal-2024-notion} propose that the complexity of theory of mind tasks can be characterized by the number of states an observer must track. This view presupposes that a task comes with a natural partition into discrete states. We argue that such partitions exist only under specific structural conditions. 

\begin{wrapfigure}{r}{0.5\textwidth}
\centering
\vspace{-4pt}
\includegraphics[width=\linewidth, trim=10cm 1cm 10cm 5cm, clip]{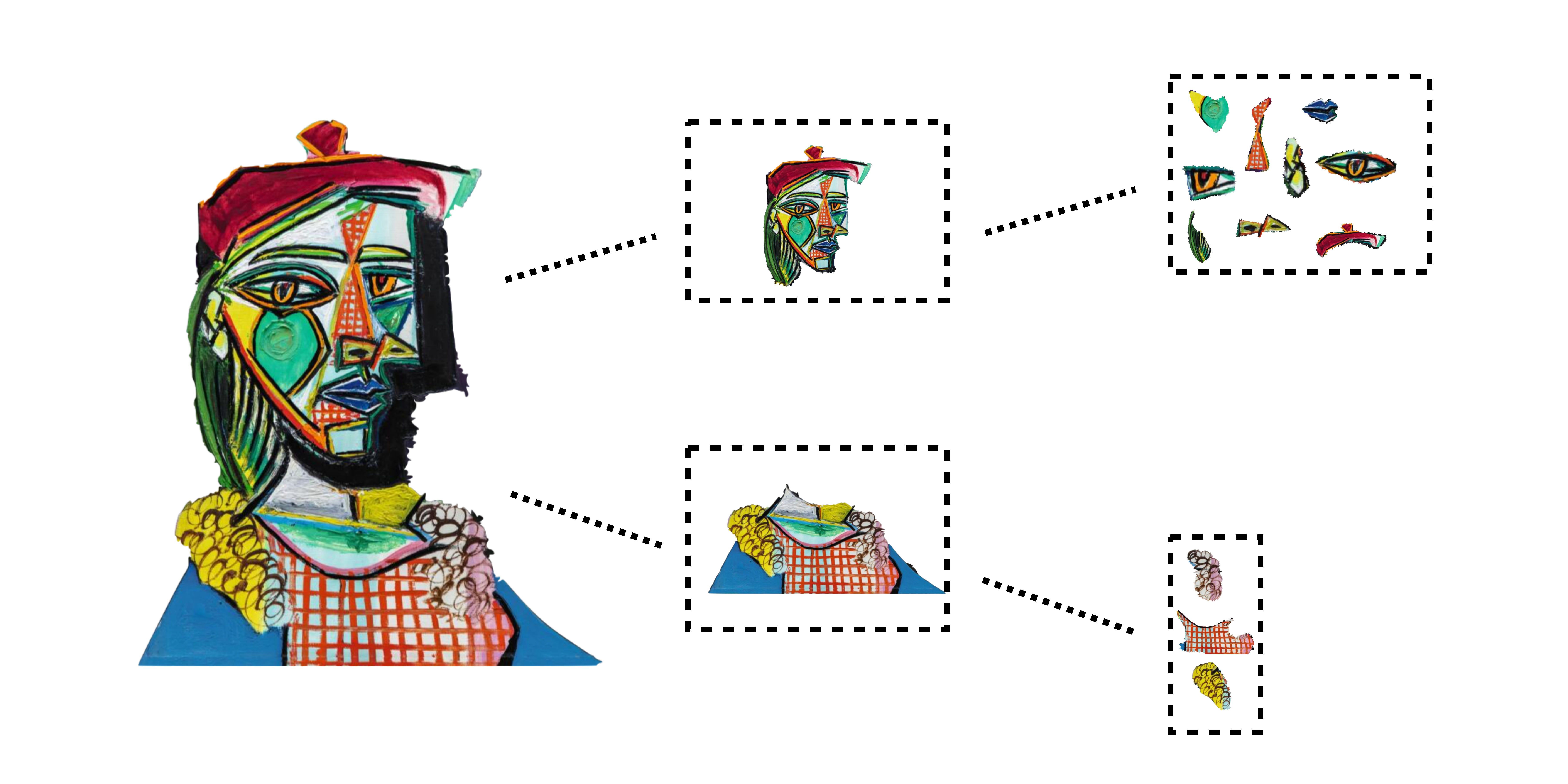}
\caption{The Picasso Thesis}
\label{fig:tommm}
\vspace{-8pt}
\end{wrapfigure}

We illustrates our position with a thought experiment: consider Picasso's \emph{Lover in a Beret} in Figure~\ref{fig:tommm}. If asked to interpret Picasso's intentions, a typical human viewer may first recognize a human face on the painting. This recognition constitutes immediate progress towards attributing Picasso's intention, after which the viewer can further decompose the face into sub-components (e.g., nose, eyes). Now imagine presenting the same painting to an alien with human-comparable intelligence but whose evolution yields no similarity to any face-ful being. If the alien has never encountered a face, it is unreasonable to expect even the shallow interpretations that humans achieve effortlessly. The alien is likely to form a partition of the painting starkly different from Figure~\ref{fig:tommm}, relying more on geometric primitives. For humans, the painting has fewer states to track because the viewer's equipment of contingent knowledge compresses multiple geometric objects into a single token. This thought experiment exposes a limit for all state-counting perspectives on ToM complexity: the number of states depends on the observer, and the observer is restricted by the long tail of contingent knowledge\citep{badhe2026longtailknowledgelargelanguage}. Complexity derived from state statistics is thus observer-relative and cannot be treated as an invariant property across ToM tasks. Instead, the partitioning procedure that produces those states and its upstream epistemological commitments become the central object of analysis. \textbf{Any benchmark that exhibits a shallow separation between state tracking and mentalistic reasoning eliminates this relativity by fixing the semantics of the vocabulary relative to which states are tokenized.}

Fixing a vocabulary and its update rules effectively encodes an algorithm. The aforementioned subset of benchmarks specifies, through its verb inventory and the semantics of those verbs, a procedure for computing state transitions. Under such designs, ToM evaluation becomes a form of algorithm recovery. The extent to which world knowledge participates in this recovery admits a hierarchy. At the minimal end, the algorithm is recoverable from surface forms alone. For example, ''Mary moves the ball to the basket'' encodes agent, object, and destination in its surface form, ready to be parsed. Sally--Anne benchmarks operate near this regime. One step beyond syntax, many subjective predicates implicitly parameterize a continuous, sometimes probabilistic internal degree whose value is mediated by context beyond surface forms. For instance, to properly compute \emph{hide}-related conditions, one may need to track an implicit variable capturing an agent's aversion to being discovered. Including such predicates increases benchmark expressivity. Further along the hierarchy, recovery requires compositional integration of long-tail world knowledge embedded in narrative form. A character's reaction may depend on biological inheritance, professional norms, or cultural and historical conventions. Computing the correct state transition then requires recognizing the relevance of such facts and retrieving them from outside the text. The hierarchy suggests two directions for benchmark evolution. First, benchmarks can vary the algorithm associated with the same verb across instances. Second, benchmarks can expand their world model to align with systems of human-level integration of knowledge. BigToM \citep{gandhi2023understanding} is one example, where language models generate novel narratives using long-tail knowledge. However, BigToM is now saturated, with Gemini 2.0 scoring 86.33\% and GPT-4o scoring 82.42\% \citep{kim2025tracing}. In benchmarks of this kind, only a system that has acquired broad world knowledge (a language model or a human) can possibly recognize the relevance of such knowledge and retrieve it appropriately. 

\subsection*{AI use statement}

In this work, we used generative AI tools (LLM coding assistants) to help implement
methods, specifically parts of the data-processing and evaluation code, and to generate synthetic
PDDL data. 
We did not use generative AI tools to propose hypotheses, create philosophical implication,
formulate or prove mathematical claims, or interpret qualitative results; translation data analysis are not applicable to this work.
Additionally, we used generative AI tools to format tables and figures, and edit the paper for readability.
We have reviewed all AI-assisted work: all AI-assisted code was checked by the authors and validated against held-out
evaluations, and every reported number was checked against the logged result files.
We take responsibility for the final content of this work, including text, claims
and artifacts produced with the aid of generative AI.

\bibliography{custom}

\begin{thebibliography}{27}
\providecommand{\natexlab}[1]{#1}
\providecommand{\url}[1]{\texttt{#1}}
\expandafter\ifx\csname urlstyle\endcsname\relax
  \providecommand{\doi}[1]{doi: #1}\else
  \providecommand{\doi}{doi: \begingroup \urlstyle{rm}\Url}\fi

\bibitem[Agrawal et~al.(2026)Agrawal, Tan, Soylu, Ziems, Khare, Opsahl-Ong,
  Singhvi, Shandilya, Ryan, Jiang, Potts, Sen, Dimakis, Stoica, Klein, Zaharia,
  and Khattab]{agrawal2025gepa}
Lakshya~A Agrawal, Shangyin Tan, Dilara Soylu, Noah Ziems, Rishi Khare, Krista
  Opsahl-Ong, Arnav Singhvi, Herumb Shandilya, Michael~J Ryan, Meng Jiang,
  Christopher Potts, Koushik Sen, Alex Dimakis, Ion Stoica, Dan Klein, Matei
  Zaharia, and Omar Khattab.
\newblock {GEPA}: Reflective prompt evolution can outperform reinforcement
  learning.
\newblock In \emph{The Fourteenth International Conference on Learning
  Representations}, 2026.
\newblock URL \url{https://openreview.net/forum?id=RQm2KQTM5r}.

\bibitem[Badhe et~al.(2026)Badhe, Shah, and
  Kathrotia]{badhe2026longtailknowledgelargelanguage}
Sanket Badhe, Deep Shah, and Nehal Kathrotia.
\newblock Long-tail knowledge in large language models: Taxonomy, mechanisms,
  interventions and implications, 2026.
\newblock URL \url{https://arxiv.org/abs/2602.16201}.

\bibitem[Baron-Cohen et~al.(1985)Baron-Cohen, Leslie, and
  Frith]{baroncohen1985does}
Simon Baron-Cohen, Alan~M. Leslie, and Uta Frith.
\newblock Does the autistic child have a ``theory of mind''?
\newblock \emph{Cognition}, 21\penalty0 (1):\penalty0 37--46, 1985.
\newblock \doi{10.1016/0010-0277(85)90022-8}.
\newblock URL \url{https://doi.org/10.1016/0010-0277(85)90022-8}.

\bibitem[Chan et~al.(2024)Chan, Jiayang, Yim, Deng, Fan, Li, Liu, Zhang, Wang,
  and Song]{chan-etal-2024-negotiationtom}
Chunkit Chan, Cheng Jiayang, Yauwai Yim, Zheye Deng, Wei Fan, Haoran Li, Xin
  Liu, Hongming Zhang, Weiqi Wang, and Yangqiu Song.
\newblock {N}egotiation{T}o{M}: A benchmark for stress-testing machine theory
  of mind on negotiation surrounding.
\newblock In Yaser Al-Onaizan, Mohit Bansal, and Yun-Nung Chen (eds.),
  \emph{Findings of the Association for Computational Linguistics: EMNLP 2024},
  pp.\  4211--4241, Miami, Florida, USA, November 2024. Association for
  Computational Linguistics.
\newblock \doi{10.18653/v1/2024.findings-emnlp.244}.
\newblock URL \url{https://aclanthology.org/2024.findings-emnlp.244/}.

\bibitem[Gandhi et~al.(2023)Gandhi, Fr{\"a}nken, Gerstenberg, and
  Goodman]{gandhi2023understanding}
Kanishk Gandhi, Jan-Philipp Fr{\"a}nken, Tobias Gerstenberg, and Noah Goodman.
\newblock Understanding social reasoning in language models with language
  models.
\newblock In \emph{Thirty-seventh Conference on Neural Information Processing
  Systems Datasets and Benchmarks Track}, 2023.
\newblock URL \url{https://openreview.net/forum?id=8bqjirgxQM}.

\bibitem[Ghallab et~al.(1998)Ghallab, Howe, Knoblock, McDermott, Ram, Veloso,
  Weld, and Wilkins]{aeronautiques1998pddl}
Malik Ghallab, Adele Howe, Craig Knoblock, Drew McDermott, Ashwin Ram, Manuela
  Veloso, Daniel Weld, and David Wilkins.
\newblock {PDDL} - the planning domain definition language.
\newblock \emph{Technical Report, Tech. Rep.}, 1998.

\bibitem[Hou et~al.(2024)Hou, Zhang, Shen, Wu, and Lu]{hou-etal-2024-timetom}
Guiyang Hou, Wenqi Zhang, Yongliang Shen, Linjuan Wu, and Weiming Lu.
\newblock {T}ime{T}o{M}: Temporal space is the key to unlocking the door of
  large language models' theory-of-mind.
\newblock In Lun-Wei Ku, Andre Martins, and Vivek Srikumar (eds.),
  \emph{Findings of the Association for Computational Linguistics: ACL 2024},
  pp.\  11532--11547, Bangkok, Thailand, August 2024. Association for
  Computational Linguistics.
\newblock \doi{10.18653/v1/2024.findings-acl.685}.
\newblock URL \url{https://aclanthology.org/2024.findings-acl.685/}.

\bibitem[Huang et~al.(2024)Huang, La~Malfa, Marro, Asperti, Cohn, and
  Wooldridge]{huang-etal-2024-notion}
X.~Angelo Huang, Emanuele La~Malfa, Samuele Marro, Andrea Asperti, Anthony~G.
  Cohn, and Michael~J. Wooldridge.
\newblock A notion of complexity for theory of mind via discrete world models.
\newblock In Yaser Al-Onaizan, Mohit Bansal, and Yun-Nung Chen (eds.),
  \emph{Findings of the Association for Computational Linguistics: EMNLP 2024},
  pp.\  2964--2983, Miami, Florida, USA, November 2024. Association for
  Computational Linguistics.
\newblock \doi{10.18653/v1/2024.findings-emnlp.167}.
\newblock URL \url{https://aclanthology.org/2024.findings-emnlp.167/}.

\bibitem[Jin et~al.(2024)Jin, Wu, Cao, Xiang, Kuo, Hu, Ullman, Torralba,
  Tenenbaum, and Shu]{jin2024mmtom}
Chuanyang Jin, Yutong Wu, Jing Cao, Jiannan Xiang, Yen-Ling Kuo, Zhiting Hu,
  Tomer Ullman, Antonio Torralba, Joshua Tenenbaum, and Tianmin Shu.
\newblock {MMT}o{M}-{QA}: Multimodal theory of mind question answering.
\newblock In Lun-Wei Ku, Andre Martins, and Vivek Srikumar (eds.),
  \emph{Proceedings of the 62nd Annual Meeting of the Association for
  Computational Linguistics (Volume 1: Long Papers)}, pp.\  16077--16102,
  Bangkok, Thailand, August 2024. Association for Computational Linguistics.
\newblock \doi{10.18653/v1/2024.acl-long.851}.
\newblock URL \url{https://aclanthology.org/2024.acl-long.851/}.

\bibitem[Kim et~al.(2023)Kim, Sclar, Zhou, Bras, Kim, Choi, and
  Sap]{kim2023fantom}
Hyunwoo Kim, Melanie Sclar, Xuhui Zhou, Ronan~Le Bras, Gunhee Kim, Yejin Choi,
  and Maarten Sap.
\newblock {FANT}om: A benchmark for stress-testing machine theory of mind in
  interactions.
\newblock In \emph{The 2023 Conference on Empirical Methods in Natural Language
  Processing}, 2023.
\newblock URL \url{https://openreview.net/forum?id=5TEfD2GBUc}.

\bibitem[Kim et~al.(2025)Kim, Sclar, Zhi-Xuan, Ying, Levine, Liu, Tenenbaum,
  and Choi]{kim2025tracing}
Hyunwoo Kim, Melanie Sclar, Tan Zhi-Xuan, Lance Ying, Sydney Levine, Yang Liu,
  Joshua~B. Tenenbaum, and Yejin Choi.
\newblock Hypothesis-driven theory-of-mind reasoning for large language models.
\newblock In \emph{Second Conference on Language Modeling}, 2025.
\newblock URL \url{https://openreview.net/forum?id=yGQqTuSJPK}.

\bibitem[Kosinski(2024)]{kosinski2024evaluating}
Michal Kosinski.
\newblock Evaluating large language models in theory of mind tasks.
\newblock \emph{Proceedings of the National Academy of Sciences}, 121\penalty0
  (45):\penalty0 e2405460121, 2024.
\newblock \doi{10.1073/pnas.2405460121}.
\newblock URL \url{https://www.pnas.org/doi/abs/10.1073/pnas.2405460121}.

\bibitem[Le et~al.(2019)Le, Boureau, and Nickel]{le-etal-2019-revisiting-tomi}
Matthew Le, Y-Lan Boureau, and Maximilian Nickel.
\newblock Revisiting the evaluation of theory of mind through question
  answering.
\newblock In Kentaro Inui, Jing Jiang, Vincent Ng, and Xiaojun Wan (eds.),
  \emph{Proceedings of the 2019 Conference on Empirical Methods in Natural
  Language Processing and the 9th International Joint Conference on Natural
  Language Processing (EMNLP-IJCNLP)}, pp.\  5872--5877, Hong Kong, China,
  November 2019. Association for Computational Linguistics.
\newblock \doi{10.18653/v1/D19-1598}.
\newblock URL \url{https://aclanthology.org/D19-1598/}.

\bibitem[Liu et~al.(2023)Liu, Jiang, Zhang, Liu, Zhang, Biswas, and
  Stone]{llmp}
Bo~Liu, Yuqian Jiang, Xiaohan Zhang, Qiang Liu, Shiqi Zhang, Joydeep Biswas,
  and Peter Stone.
\newblock Llm+p: Empowering large language models with optimal planning
  proficiency, 2023.
\newblock URL \url{https://arxiv.org/abs/2304.11477}.

\bibitem[Ma et~al.(2023)Ma, Gao, and Xu]{ma2023tomchallenges}
Xiaomeng Ma, Lingyu Gao, and Qihui Xu.
\newblock {T}o{MC}hallenges: A principle-guided dataset and diverse evaluation
  tasks for exploring theory of mind.
\newblock In Jing Jiang, David Reitter, and Shumin Deng (eds.),
  \emph{Proceedings of the 27th Conference on Computational Natural Language
  Learning (CoNLL)}, pp.\  15--26, Singapore, December 2023. Association for
  Computational Linguistics.
\newblock \doi{10.18653/v1/2023.conll-1.2}.
\newblock URL \url{https://aclanthology.org/2023.conll-1.2/}.

\bibitem[OpenAI(2024)]{openai2024gpt4ocard}
OpenAI.
\newblock Gpt-4o system card, 2024.
\newblock URL \url{https://arxiv.org/abs/2410.21276}.

\bibitem[Sclar et~al.(2023)Sclar, Kumar, West, Suhr, Choi, and
  Tsvetkov]{sclar-etal-2023-minding-symbolictom}
Melanie Sclar, Sachin Kumar, Peter West, Alane Suhr, Yejin Choi, and Yulia
  Tsvetkov.
\newblock Minding language models' (lack of) theory of mind: A plug-and-play
  multi-character belief tracker.
\newblock In Anna Rogers, Jordan Boyd-Graber, and Naoaki Okazaki (eds.),
  \emph{Proceedings of the 61st Annual Meeting of the Association for
  Computational Linguistics (Volume 1: Long Papers)}, pp.\  13960--13980,
  Toronto, Canada, July 2023. Association for Computational Linguistics.
\newblock \doi{10.18653/v1/2023.acl-long.780}.
\newblock URL \url{https://aclanthology.org/2023.acl-long.780/}.

\bibitem[Shi et~al.(2025)Shi, Ye, Fang, Jin, Isik, Kuo, and
  Shu]{shi2024mumatommultimodalmultiagenttheory}
Haojun Shi, Suyu Ye, Xinyu Fang, Chuanyang Jin, Leyla Isik, Yen-Ling Kuo, and
  Tianmin Shu.
\newblock Muma-tom: Multi-modal multi-agent theory of mind, 2025.
\newblock URL \url{https://arxiv.org/abs/2408.12574}.

\bibitem[Ullman(2023)]{ullman2023large}
Tomer Ullman.
\newblock Large language models fail on trivial alterations to theory-of-mind
  tasks, 2023.
\newblock URL \url{https://arxiv.org/abs/2302.08399}.

\bibitem[Wilf et~al.(2024)Wilf, Lee, Liang, and
  Morency]{wilf-etal-2024-think-simtom}
Alex Wilf, Sihyun Lee, Paul~Pu Liang, and Louis-Philippe Morency.
\newblock Think twice: Perspective-taking improves large language models'
  theory-of-mind capabilities.
\newblock In Lun-Wei Ku, Andre Martins, and Vivek Srikumar (eds.),
  \emph{Proceedings of the 62nd Annual Meeting of the Association for
  Computational Linguistics (Volume 1: Long Papers)}, pp.\  8292--8308,
  Bangkok, Thailand, August 2024. Association for Computational Linguistics.
\newblock \doi{10.18653/v1/2024.acl-long.451}.
\newblock URL \url{https://aclanthology.org/2024.acl-long.451/}.

\bibitem[Wimmer \& Perner(1983)Wimmer and Perner]{wimmer1983beliefs}
Heinz Wimmer and Josef Perner.
\newblock Beliefs about beliefs: Representation and constraining function of
  wrong beliefs in young children's understanding of deception.
\newblock \emph{Cognition}, 13\penalty0 (1):\penalty0 103--128, 1983.
\newblock \doi{10.1016/0010-0277(83)90004-5}.
\newblock URL \url{https://doi.org/10.1016/0010-0277(83)90004-5}.

\bibitem[Wu et~al.(2023)Wu, He, Jia, Mihalcea, Chen, and
  Deng]{wu-etal-2023-hitom}
Yufan Wu, Yinghui He, Yilin Jia, Rada Mihalcea, Yulong Chen, and Naihao Deng.
\newblock Hi-{T}o{M}: A benchmark for evaluating higher-order theory of mind
  reasoning in large language models.
\newblock In Houda Bouamor, Juan Pino, and Kalika Bali (eds.), \emph{Findings
  of the Association for Computational Linguistics: EMNLP 2023}, pp.\
  10691--10706, Singapore, December 2023. Association for Computational
  Linguistics.
\newblock \doi{10.18653/v1/2023.findings-emnlp.717}.
\newblock URL \url{https://aclanthology.org/2023.findings-emnlp.717/}.

\bibitem[Xu et~al.(2024)Xu, Zhao, Zhu, Du, and He]{xu-etal-2024-opentom}
Hainiu Xu, Runcong Zhao, Lixing Zhu, Jinhua Du, and Yulan He.
\newblock {O}pen{T}o{M}: A comprehensive benchmark for evaluating
  theory-of-mind reasoning capabilities of large language models.
\newblock In Lun-Wei Ku, Andre Martins, and Vivek Srikumar (eds.),
  \emph{Proceedings of the 62nd Annual Meeting of the Association for
  Computational Linguistics (Volume 1: Long Papers)}, pp.\  8593--8623,
  Bangkok, Thailand, August 2024. Association for Computational Linguistics.
\newblock \doi{10.18653/v1/2024.acl-long.466}.
\newblock URL \url{https://aclanthology.org/2024.acl-long.466/}.

\bibitem[Xu et~al.(2025)Xu, Qi, Li, Zhou, Du, Catmur, and
  He]{xu-etal-2025-enigmatom}
Hainiu Xu, Siya Qi, Jiazheng Li, Yuxiang Zhou, Jinhua Du, Caroline Catmur, and
  Yulan He.
\newblock {E}nigma{T}o{M}: Improve {LLM}s' theory-of-mind reasoning
  capabilities with neural knowledge base of entity states.
\newblock In Wanxiang Che, Joyce Nabende, Ekaterina Shutova, and Mohammad~Taher
  Pilehvar (eds.), \emph{Findings of the Association for Computational
  Linguistics: ACL 2025}, pp.\  13598--13622, Vienna, Austria, July 2025.
  Association for Computational Linguistics.
\newblock ISBN 979-8-89176-256-5.
\newblock \doi{10.18653/v1/2025.findings-acl.699}.
\newblock URL \url{https://aclanthology.org/2025.findings-acl.699/}.

\bibitem[Zhang et~al.(2025)Zhang, Jin, Jia, Zhang, and Shu]{zhang2025autotom}
Zhining Zhang, Chuanyang Jin, Mung~Yao Jia, Shunchi Zhang, and Tianmin Shu.
\newblock Autotom: Scaling model-based mental inference via automated agent
  modeling.
\newblock In \emph{The Thirty-ninth Annual Conference on Neural Information
  Processing Systems}, 2025.
\newblock URL \url{https://openreview.net/forum?id=oeZZusZheP}.

\bibitem[Zhou et~al.(2026)Zhou, Liu, Yerukola, Kim, and
  Sap]{zhou2026socialworldmodels}
Xuhui Zhou, Jiarui Liu, Akhila Yerukola, Hyunwoo Kim, and Maarten Sap.
\newblock Social world models.
\newblock In \emph{Third Conference on Language Modeling}, 2026.
\newblock URL \url{https://openreview.net/forum?id=JGSDyABvPH}.

\bibitem[Zhu et~al.(2025)Zhu, Singh, Jia, and
  Thomason]{zhu-etal-2025-language-models}
Wang~Bill Zhu, Ishika Singh, Robin Jia, and Jesse Thomason.
\newblock Language models can infer action semantics for symbolic planners from
  environment feedback.
\newblock In Luis Chiruzzo, Alan Ritter, and Lu~Wang (eds.), \emph{Proceedings
  of the 2025 Conference of the Nations of the Americas Chapter of the
  Association for Computational Linguistics: Human Language Technologies
  (Volume 1: Long Papers)}, pp.\  8751--8773, Albuquerque, New Mexico, April
  2025. Association for Computational Linguistics.
\newblock ISBN 979-8-89176-189-6.
\newblock \doi{10.18653/v1/2025.naacl-long.440}.
\newblock URL \url{https://aclanthology.org/2025.naacl-long.440/}.

\end{thebibliography}
\bibliographystyle{iclr2026_conference}

\clearpage

\appendix
\section*{Appendix}
\startcontents[appendix]
\renewcommand{\thesection}{\Alph{section}}
\printcontents[appendix]{}{1}{\setcounter{tocdepth}{3}}
\setcounter{section}{0}

\section{Prompts}
\label{appsec:prompt}

We list the PDDL domain file (Listing~\ref{lst:pddl}) and the prompts used for LLM parsing and ToM question answering (Listings~\ref{lst:prompt_pf}, \ref{lst:prompt_action}, and \ref{lst:prompt_tomqa}).
The domain file includes a small custom simulator for effects that are convenient to write as ``forall'' updates but are not supported by the PDDL executor used in our implementation.

\lstdefinelanguage{PDDL}{
  keywords={define,domain,requirements,types,predicates,action, parameters,precondition,effect,and,or,not,forall,when},
  keywordstyle=\color{blue}\bfseries,
  comment=[l]{;},
  commentstyle=\color{gray},
  basicstyle=\ttfamily\small,
  columns=fullflexible,
  keepspaces=true,
  showstringspaces=false,
  breaklines=true,
  frame=single
}

\begin{lstlisting}[language=PDDL, caption={PDDL domain used in \ourmethod}, label={lst:pddl}]
(define (domain tom_general)
  (:requirements :strips)
  (:types agent obj loc utterance)    ; agent, object, location and utterance

  (:predicates
    ;; Required for all scenarios, if for communication data, create an abstract location (e.g. conversation) for everyone present
    (at ?a - agent ?l - loc)           ; agent a is at location l.

    ;; Physical movement, ONLY use when the data is about MOVING agents
    (is-open ?c)                       ; c is open, c is a container, which can be an object or a location.
    (is-loc ?c)                        ; container c is a location (e.g. a room).
    (in ?o - obj ?c)                   ; object o is in container c, which can be an object or a location, do NOT declare object X on some other object Y, instead, declare object X "in" the same container as object Y.
    (holding ?a - agent ?o - obj)      ; agent is holding object o.
    (seen ?a - agent ?o - obj ?l - loc?)         ; agent a has seen object o at location l.

    ;; Communication tracking, ONLY use when agents have active conversations
    (said ?a - agent ?u - utterance)    ; agent a said utterance u.
    (heard ?a - agent ?u - utterance)   ; agent a heard utterance u.
  )

  ;; Physical movement, ONLY use when the data is about MOVING agents
  ;; agent moves from a start room to a dest room, sees all objects visible in the destination room
  (:action move
    :parameters (?a - agent ?start - loc ?dest - loc)
    :precondition (at ?a ?start)
    :effect (and
      (forall (?o - obj)
        when (in ?o ?dest)
        (seen ?a ?o ?dest)
      )
    )
  )

  ;; Physical movement, ONLY use when the data is about MOVING agents
  ;; open a container c at location l, the container can be an object or a location.
  (:action open
    :parameters  (?a - agent ?c ?l - loc)
    :precondition (and
      (at ?a ?l)
      (or
        (in ?c ?l)
        (is-loc ?c)
      )
    )
    :effect (and
      (is-open ?c)
      (forall (?o - obj)
        when (in ?o ?c)
        (seen ?a ?o ?l)
      )
    )
  )

  ;; Physical movement, ONLY use when the data is about MOVING agents
  ;; close a container c at location l, the container can be an object or a location.
  (:action close
    :parameters (?a - agent ?c ?l - loc)
    :precondition (and
      (at ?a ?l)
      (is-open ?c)
      (or
        (in ?c ?l)
        (is-loc ?c)
      )
    )
    :effect (not (is-open ?c))
  )

  ;; Physical movement, ONLY use when the data is about MOVING agents
  ;; grab an object o at location l
  (:action grab
    :parameters (?a - agent ?o - obj ?l - loc)
    :precondition (and
      (at ?a ?l)
      (in ?o ?l)
      (not (holding ?a ?o))
    )
    :effect (and
      (holding ?a ?o)
      (not (in ?o ?l))
      (seen ?a ?o ?l)
    )
  )

  ;; Physical movement, ONLY use when the data is about MOVING agents
  ;; drop an object o at location l
  (:action drop
    :parameters (?a - agent ?o - obj ?l - loc)
    :precondition (and
      (at ?a ?l)
      (holding ?a ?o)
    )
    :effect (and
      (not (holding ?a ?o))
      (in ?o ?l)
    )
  )

  ;; Communication tracking, ONLY use when agents have active conversations
  ;; agent a tells others an utterance u
  (:action tell
    :parameters (?a - agent ?u - utterance)
    :precondition ()
    :effect (and
      (said ?a ?u)
      ;; Hear of other agents will be handled in simulation
    )
  )

  ;; Communication tracking, ONLY use when agents have active conversations
  ;; agent hears an utterance u at location l (while present), can be triggered by action tell
  (:action hear
    :parameters (?a - agent ?l - loc ?u - utterance)
    :precondition (and (at ?a ?l))
    :effect (and (heard ?a ?u))
  )
)
\end{lstlisting}

\lstdefinelanguage{Prompt}{
  basicstyle=\ttfamily\small,
  columns=fullflexible,
  keepspaces=true,
  showstringspaces=false,
  breaklines=true,
  frame=single
}

\begin{lstlisting}[language=Prompt,
                   caption={Prompt template used for PDDL problem file generation},
                   label={lst:prompt_pf}]
The domain types and predicates definition for the current environment:
[[domain_type_pred]]
Given the above domain types and predicates, please write the PDDL problem file for the following environment. 
All rooms (locations) are seen.
All objects are NOT seen.
Please list the objects (with their types) and the initial state faithfully, but only a dummy goal state like (:goal ).

Environment:
[[environment]]
\end{lstlisting}

\begin{lstlisting}[language=Prompt,
                   caption={Prompt template used for PDDL action extraction},
                   label={lst:prompt_action}]
Domain:
[[domain]]

Initially, the environment states are
[[init_state]]

Parse the following actions including the ongoing actions into PDDL format given the domain PDDL. 
Only take the single-agent perspective actions: move, open and close in the domain PDDL. 
Only use the objects provided. 
Format the final PDDL action list as
"Executed Actions:
(action1 ...)
...
Ongoing Actions:
(action2 ...)
..."

[[action]]
\end{lstlisting}

\begin{lstlisting}[language=Prompt,
                   caption={Prompt template used for ToM question answering},
                   label={lst:prompt_tomqa}]
[[names]] was/were in an environment with these actions allowed: [[domain_action_name]]. [[action]]

We break down the actions taken by [[names]] with location, the seen objects lists, and heard utterance lists.
[[act_state_seq]][[ongoing_prompt]]

[[question]]
[[notes]]

Please answer based on the above information.
[[final_format]]
\end{lstlisting}

\section{Experiment Details}
\label{appsec:details}

\subsection{Implementation Details}
\label{appsec:impdetails}

For all models and all benchmarks, we use temperature $0$ and a maximum generation length of 4,096 tokens for LLM inference.
To avoid data leakage, we stratify 50 examples per sub-category on each benchmark and exclude the examples used to design the PDDL domain file and prompt templates.

\subsection{Error Analysis}
\label{appsec:error}

\begin{lstlisting}[language=Prompt,
                   caption={Example on failed example in \mmtom with reasoning errors.},
                   label={lst:fail_reasoning}]
Narratives:
What's inside the apartment: The apartment consists of a bedroom, a bathroom, a living room, and a kitchen. 
In the bedroom, there is a coffee table with a dish bowl, a remote control, a plate, and a water glass on it.
The bathroom houses a cabinet, which is currently empty.
The living room is furnished with a cabinet, a coffee table, a sofa, and a desk. The cabinet contains three apples, a condiment bottle, three water glasses, a remote control, a wine glass, and a dish bowl. A water glass is placed on the coffee table. 
The kitchen is equipped with a fridge, a stove, a kitchen table, and a microwave. Inside the fridge, there are two dish bowls and a bottle of wine. The stove contains four plates, a salmon, and a cupcake. The microwave holds a salmon, a condiment bottle, and a bag of chips. 
Actions taken by Karen: Karen is situated in the kitchen. She strides towards the refrigerator, opens it, and subsequently closes it. Following this, she opens the microwave and then securely shuts it. 

-------------------------------------------------

Predicted action sequence:
(move Karen kitchen fridge)
(open Karen fridge kitchen)
(close Karen fridge kitchen)
(open Karen microwave kitchen)
(close Karen microwave kitchen)

-------------------------------------------------

Verified action sequence:
Initially, Karen is situated in the kitchen 
Seen objects: fridge, kitchen_table, microwave
Heard utterance: N/A

1. (open Karen fridge kitchen) 
Seen objects: fridge, kitchen_table, microwave, dish_bowl_3, dish_bowl_4, wine_bottle
Heard utterance: N/A

2. (close Karen fridge kitchen)
Seen objects: fridge, kitchen_table, microwave, dish_bowl_3, dish_bowl_4, wine_bottle
Heard utterance: N/A

3. (open Karen microwave kitchen)
Seen objects: fridge, kitchen_table, microwave, dish_bowl_3, dish_bowl_4, wine_bottle, salmon_2, condiment_bottle_2, chips.
Heard utterance: N/A

4. (close Karen microwave kitchen)
Seen objects: fridge, kitchen_table, microwave, dish_bowl_3, dish_bowl_4, wine_bottle, salmon_2, condiment_bottle_2, chips.
Heard utterance: N/A

-------------------------------------------------

Question:
Which one of the following statements is more likely to be true? (a) Karen has been trying to get a bag of chips. (b) Karen has been trying to get a remote control.

-------------------------------------------------

Reasoning & Answer:
Karen interacted with the fridge and the microwave, both of which revealed the presence of chips upon opening the microwave. Since she did not pick up the chips after seeing them, it suggests she might be looking for something else. However, she did not interact with any other objects or containers that might suggest a search for a remote control. Therefore, the presence of chips and the lack of any interaction with objects related to a remote control make option (a) more likely. The answer is a.
\end{lstlisting}

\begin{lstlisting}[language=Prompt,
                   caption={Example on failed example in \mmtom with parsing errors.},
                   label={lst:fail_parsing}]
Narratives:
What's inside the apartment: The apartment consists of a bedroom, kitchen, living room, and bathroom. 
In the bedroom, there is a coffee table and a desk
The kitchen is equipped with four cabinets, a fridge, a microwave, a kitchen table, and a stove. The fourth cabinet from the left houses two water glasses, while the second and third cabinets are empty. The first cabinet contains a water glass, a dish bowl, and a condiment bottle. Inside the fridge, there is a plate and an apple, and a cupcake is in the microwave. The stove contains a plate and a piece of salmon. 
The living room features a cabinet, a coffee table, a sofa, and a desk. The cabinet is filled with two books, a water glass, a bag of chips, a bottle of wine, two remote controls, two apples, a wine glass, and a condiment bottle. The coffee table holds two dish bowls and a water glass. 
The bathroom has a cabinet, which is currently empty.
Actions taken by David: David is situated in the kitchen. He strides towards the fourth kitchen cabinet, opens it, subsequently closes it, and is now preparing to open the third kitchen cabinet. 


-------------------------------------------------

Predicted action sequence:
(move David kitchen cabinet_kitchen_4) 
(open David cabinet_kitchen_4 kitchen) 
(close David cabinet_kitchen_4 kitchen)

Ongoing action sequence:
(open David cabinet_kitchen_3 kitchen)


-------------------------------------------------

Verified action sequence:
Initially, David is situated in the kitchen 
Seen objects: stove
Heard utterance: N/A

1. (move David kitchen cabinet_kitchen_4) 
Seen objects: stove, water_glass_1, water_glass_2
Heard utterance: N/A

[[Note that cabinet_kitchen_4  is parsed as a location with (is-loc cabinet_kitchen_4), making the action sequence stops at 'move', also cabinet_kitchen_1, cabinet_kitchen_2, cabinet_kitchen_3, cabinet_kitchen_4 are not listed are seen objects due to this.]]


-------------------------------------------------

Question:
Which one of the following statements is more likely to be true? (a) Karen has been trying to get a bag of chips. (b) Karen has been trying to get a remote control.

-------------------------------------------------

Reasoning & Answer:
David is searching through the cabinets, which suggests he believes the bottle of wine might be in one of them. The answer is a.

\end{lstlisting}

\section{Ethics and Risks}
\label{appsec:checklist}

\paragraph{Potential risks.}
\ourmethod conditions LLMs on a predefined PDDL domain file that specifies permissible actions and state transitions.
While this design improves state-tracking reliability, it also means reported gains may reflect alignment with the predefined schema rather than a general increase in ToM ability.
To mitigate this risk, we release the full domain specification and prompts, and encourage future work to evaluate robustness under alternative or extended domains.

\paragraph{Data and model use.}
We use the public datasets \mmtom, \muma, and \fantom; \mmtom and \muma are released under MIT licenses.
These datasets are anonymized and contain no private human data.
GPT-4o and Claude-Sonnet-4.5 are accessible through commercial APIs.
Kimi-K2-Instruct is released under a Modified MIT License.

\paragraph{LLM use.}
We use LLMs to generate PDDL problem files and action sequences, to answer theory of mind questions in our experiments, and to improve writing fluency and grammar.

\end{document}